\documentclass[11pt,a4paper]{article}

\usepackage[utf8]{inputenc}
\usepackage[T1]{fontenc}
\usepackage{amsmath,amssymb}
\usepackage{graphicx}
\usepackage{booktabs}
\usepackage{hyperref}
\usepackage{natbib}
\usepackage{xcolor}
\usepackage{microtype}
\usepackage[margin=2.5cm]{geometry}
\usepackage{float}
\usepackage{placeins}
\usepackage[font=small,labelfont=bf]{caption}

\newcommand{\ie}{i.e.\@}

\title{MoE-Sieve: Routing-Guided LoRA for Efficient MoE Fine-Tuning}

\author{
  Andrea Manzoni \\
  \texttt{a.manzoni14@gmail.com}
}

\date{}

\begin{document}
\maketitle

% ══════════════════════════════════════════════════════════════════════
% ABSTRACT
% ══════════════════════════════════════════════════════════════════════
\begin{abstract}

Standard LoRA fine-tuning of Mixture-of-Experts (MoE) models applies adapters
to every expert, yet our profiling reveals that per-layer expert routing is
\emph{highly skewed}---a small subset of experts handles most tokens in each
layer, while many others are rarely activated (``cold''). We propose
\textbf{MoE-Sieve}, a simple routing-guided framework for LoRA
fine-tuning, and pair it with a systematic profiling study of expert
routing across architectures and tasks: (1)~profile routing counts on a
small calibration set, (2)~select the top-$k$ most-routed experts per
layer, and (3)~apply LoRA only to those experts.

Across two architecturally distinct MoE models and three diverse tasks,
tuning only the top-25\% routed experts per layer is \emph{competitive}
with full LoRA on all experts, with mean differences within
$\pm 1$~percentage point across all conditions. This comes at a
70--73\% reduction in LoRA trainable parameters, 71--73\% in adapter
checkpoint size, and up to 50\% in wall-clock training time. We also
observe a non-monotonic relationship between
expert count and seed-to-seed variance, consistent with the hypothesis
that adapting cold experts introduces gradient noise without improving
accuracy.

Further ablations show that random expert selection at matched budget
is ${\sim}$2.5~pp worse, showing that the routing signal matters, and that
greedy per-layer budget optimisation does not improve over uniform
top-$k$, yielding a practical one-line recipe: \emph{profile $\to$ count
$\to$ pick top-$k$ $\to$ fine-tune}.

\end{abstract}

% ══════════════════════════════════════════════════════════════════════
% 1. INTRODUCTION
% ══════════════════════════════════════════════════════════════════════
\section{Introduction}
\label{sec:intro}

Mixture-of-Experts (MoE) language models scale to billions of total parameters
while keeping per-token compute bounded by routing each token through a small
subset of expert modules
\citep{fedus2022switch, lepikhin2021gshard, jiang2024mixtral,
dai2024deepseekmoe, muennighoff2024olmoe}.
When fine-tuning such models with parameter-efficient methods like LoRA
\citep{hu2022lora}, the standard practice is to attach adapters to
\emph{every} expert---but not every expert contributes equally. Within any
given layer, routing is highly concentrated: a handful of ``hot'' experts
handle the bulk of tokens while many others are rarely activated and
effectively \emph{cold}. This local imbalance persists even when global
utilisation appears balanced, because the load-balancing loss used during
pre-training can equalise each expert's total workload across the full
model by letting it dominate in a \emph{different} layer, without
requiring uniform use \emph{within} any single layer.

This observation raises a natural question: \emph{if most experts in each
layer receive few tokens, why spend adapter capacity on them?}

We propose \textbf{MoE-Sieve}, a simple framework that answers this question
in three steps: (1)~run a single forward pass over the task data to count
per-layer expert activations, (2)~select the top-$k$ most-routed experts
in each layer, and (3)~apply LoRA only to those experts (plus the
always-active attention, router, and shared-expert modules). The profiling
pass adds negligible wall-clock overhead and requires no hyperparameter
search over allocation strategies.

We validate MoE-Sieve across two architecturally distinct MoE models
(OLMoE-1B-7B, Qwen1.5-MoE-A2.7B) and three tasks (Spider, GSM8K, HellaSwag),
with 8 seeds per condition. Training only the top-25\% of routed experts per
layer is competitive with full LoRA while reducing LoRA trainable parameters by
70--73\% (\S\ref{sec:results}). Before arriving at this result, we conduct a
systematic profiling study across 3 architectures and 10 datasets
(\S\ref{sec:profiling}), showing that per-layer routing skew---measured
by the coefficient of variation (CV) of expert activation counts within
each layer---is 4.0--4.9$\times$ higher than the corresponding global
CV computed across the full model. This structural property has received
limited systematic quantification in prior work.

Beyond the main comparison, we characterise a non-monotonic relationship
between expert budget and seed-to-seed variance (\S\ref{sec:beyond}),
and run ablations confirming that routing-guided selection outperforms
random at matched budget, and that greedy per-layer allocation does not
improve over uniform top-$k$ (\S\ref{sec:ablations}).

Taken together, the paper contributes both a practical selective
fine-tuning recipe and an empirical characterisation of the routing
structure and training dynamics that make such selection effective.

% ══════════════════════════════════════════════════════════════════════
% 2. RELATED WORK
% ══════════════════════════════════════════════════════════════════════
\section{Related Work}
\label{sec:related}

\paragraph{Routing dynamics and expert specialisation.}
The auxiliary load-balancing loss \citep{fedus2022switch, lepikhin2021gshard}
encourages uniform global expert utilisation, but several studies note that
per-layer routing remains skewed nonetheless.
\citet{muennighoff2024olmoe} analyse routing in OLMoE and report high expert
specialisation; \citet{dai2024deepseekmoe} introduce fine-grained expert
segmentation and shared experts to encourage stronger specialisation in
DeepSeek-MoE; Cerebras \citep{cerebras2025routerwars} visualise the phenomenon
across router types; and \citet{guo2025specialization} link uniform routing pressure to
reduced expert differentiation. Recent refinements---per-expert bias terms
\citep{deepseekv3}, loss-free balancing \citep{wang2024lossfree}, global-batch
LBL \citep{qiu2025demons}---address the \emph{training} side. We take the
complementary view: we quantify the resulting routing patterns and exploit them
for \emph{fine-tuning} efficiency.

\paragraph{Selective expert fine-tuning and pruning.}
The idea of adapting or retaining only a subset of experts appears in both
fine-tuning and compression.
ESFT \citep{wang2024esft} profiles task data through the model's existing
router to identify frequently-activated experts per layer and applies full
fine-tuning to the selected subset; our approach applies LoRA to this same
selection logic, reducing trainable parameters by a further 70--73\% while
preserving equivalence with full LoRA.
Concurrent work HELLoRA \citep{wei2025hellora} independently arrives at the
same core idea---applying LoRA to the most-routed experts per layer---and
evaluates it primarily on OLMoE-1B-7B.
Our work complements both with a systematic multi-architecture study
(3~models, 10~datasets), variance analysis linking expert budget to
seed-to-seed stability, and budget-allocation ablations. The convergence
of independent efforts on the same core idea strengthens the case
that routing-guided expert selection is a practical direction; our
multi-model experiments further suggest that the best-performing expert budget
may vary across architectures.
\citet{gao2025mola} show that higher layers benefit from more LoRA experts,
motivating layer-wise allocation; DR-LoRA \citep{drlora2025} combines routing
frequency with rank saliency.
On the compression side, MoE Pathfinder \citep{moepathfinder2024} shows that
uniform pruning across layers is suboptimal; REAP \citep{reap2025} prunes
experts using a score combining router gate values and activation norms; and
MoE-Spec \citep{moespec2025} budgets capacity using routing tails. These works confirm per-layer heterogeneity from a compression
perspective---we address the complementary fine-tuning question.

% ══════════════════════════════════════════════════════════════════════
% 3. ROUTING PROFILING
% ══════════════════════════════════════════════════════════════════════
\section{Routing Profiling: Quantifying Local Imbalance}
\label{sec:profiling}

Several studies have noted that per-layer routing is skewed in individual
MoE models (\S\ref{sec:related}), but these observations remain
qualitative and limited to single architectures. Before proposing an
expert-selection rule, we need to understand \emph{how much} routing
varies across layers, whether the pattern holds across different model
families, and whether it depends on the input task. This section
addresses these questions with a profiling study across three
architecturally diverse MoE models and ten datasets, establishing the
quantitative basis for expert selection.

\subsection{Setup and Core Finding}
\label{sec:cv}

We profile three architectures that all activate eight experts per token but
distribute them differently between shared (always-on) and routed
(gate-selected) modules:

\begin{table}[!htbp]
\centering
\footnotesize
\setlength{\tabcolsep}{5pt}
\caption{Architectural summary of the profiled MoE models. All three activate
  eight experts per token in total, but differ in how much capacity is routed
  versus shared and in expert granularity.}
\label{tab:profiling}
\begin{tabular}{lccccc}
\toprule
\textbf{Model}
  & \textbf{Layers} & \textbf{Routed} & \textbf{Shared} & \textbf{Top-$k$}
  & \textbf{Expert / FFN} \\
\midrule
OLMoE-1B-7B       & 16 & 64 & 0 & 8 & 1.0$\times$
  \\
Qwen1.5-MoE-A2.7B & 24 & 60 & 4 & 4 & 0.25$\times$
  \\
DeepSeek-MoE-16B   & 27 & 64 & 2 & 6 & 0.13$\times$
  \\
\bottomrule
\end{tabular}
\end{table}

OLMoE uses full-width experts (each expert = one complete FFN); DeepSeek and
Qwen use fine-grained experts (each expert is 1/8 or 1/4 of a dense-equivalent
FFN). This granularity difference affects routing tail behaviour
(\S\ref{sec:abl-counts-mass}) but not the core finding.

For each model we run a \emph{single gradient-free forward pass} over each of
10 calibration datasets (Spider, GSM8K, HellaSwag, ARC-Challenge, BoolQ, PIQA,
MMLU, CodeAlpaca, Wikitext, and MBPP), recording per-layer expert activation counts.

\begin{figure}[!htbp]
\centering
  \includegraphics[width=0.85\textwidth]{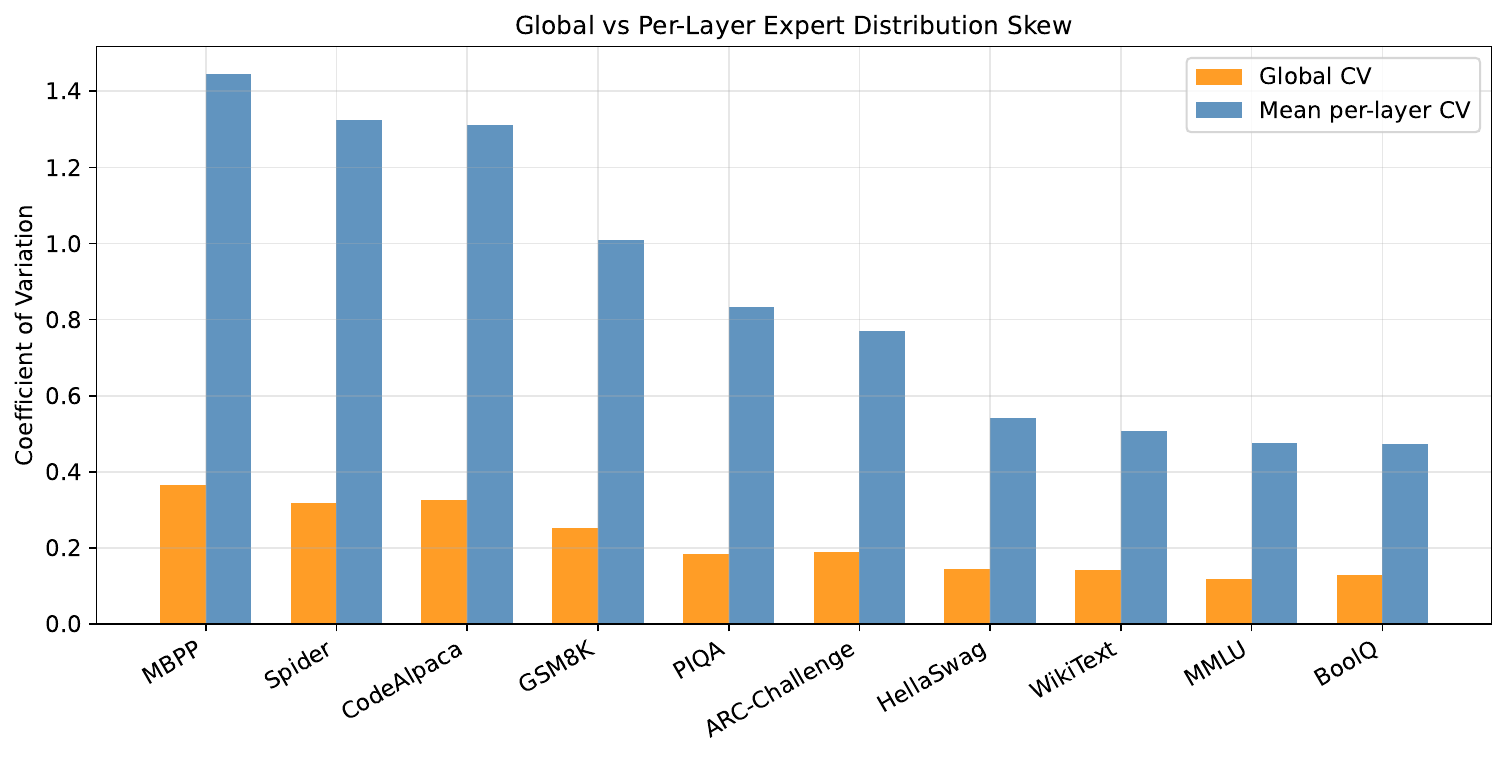}
\caption{Global vs.\ per-layer routing skew across all three models.
  Each dot is one model--dataset pair.  Per-layer CV is consistently
  4.0--4.9$\times$ higher than global CV, demonstrating the
  ``globally balanced, locally imbalanced'' phenomenon.}
\label{fig:cv-comparison}
\end{figure}

Table~\ref{tab:profiling-stats} presents the key result. We measure routing
imbalance using the coefficient of variation (CV = std/mean) of expert
activation counts: a higher CV indicates that tokens are concentrated on
fewer experts. Per-layer CV is 4.0--4.9$\times$ higher than global
CV across all three models
(Figure~\ref{fig:cv-comparison}). The load-balancing loss equalises each
expert's total workload (mean global~CV~$\leq$~0.22), but within any given layer,
routing is concentrated on a subset of experts (mean layer~CV of 0.87 for OLMoE).
This gap holds across all 30 model--dataset combinations (range: 3.6$\times$
for OLMoE~$\times$~Wikitext to 5.7$\times$ for Qwen~$\times$~GSM8K).

\begin{table}[!htbp]
\centering
\footnotesize
\setlength{\tabcolsep}{6pt}
\caption{Routing imbalance averaged over 10 datasets. \emph{Global CV}:
  coefficient of variation (standard deviation / mean) of per-expert totals
  aggregated across all layers. \emph{Layer CV}: mean within-layer CV.
  \emph{Cold\%}: fraction of experts per layer receiving $<$50\% of the
  uniform share. \emph{Cov@25\%}: fraction of token activations captured by the
  top-25\% most-routed experts per layer.}
\label{tab:profiling-stats}
\begin{tabular}{lccccc}
\toprule
\textbf{Model} & \textbf{Global CV} & \textbf{Layer CV} & \textbf{Ratio}
  & \textbf{Cold\%} & \textbf{Cov@25\%} \\
\midrule
OLMoE-1B-7B       & 0.216 & 0.869 & 4.0$\times$ & 28.5\% & 53.0\% \\
Qwen1.5-MoE-A2.7B & 0.077 & 0.374 & 4.9$\times$ & 5.2\%  & 37.4\% \\
DeepSeek-MoE-16B  & 0.106 & 0.514 & 4.9$\times$ & 11.0\% & 42.6\% \\
\bottomrule
\end{tabular}
\end{table}

\subsection{Structure of the Imbalance}
\label{sec:cross-model-profiling}

The per-layer CV numbers confirm that routing is skewed, but to design
an expert-selection strategy we need to understand the shape of that
skew in more detail. Three dimensions turn out to matter: what fraction of
token activations a small expert subset \emph{covers}, whether the identity
of hot experts changes with the input \emph{task}, and whether skew
varies with layer \emph{depth}.

\begin{figure}[!htbp]
\centering
  \includegraphics[width=\textwidth]{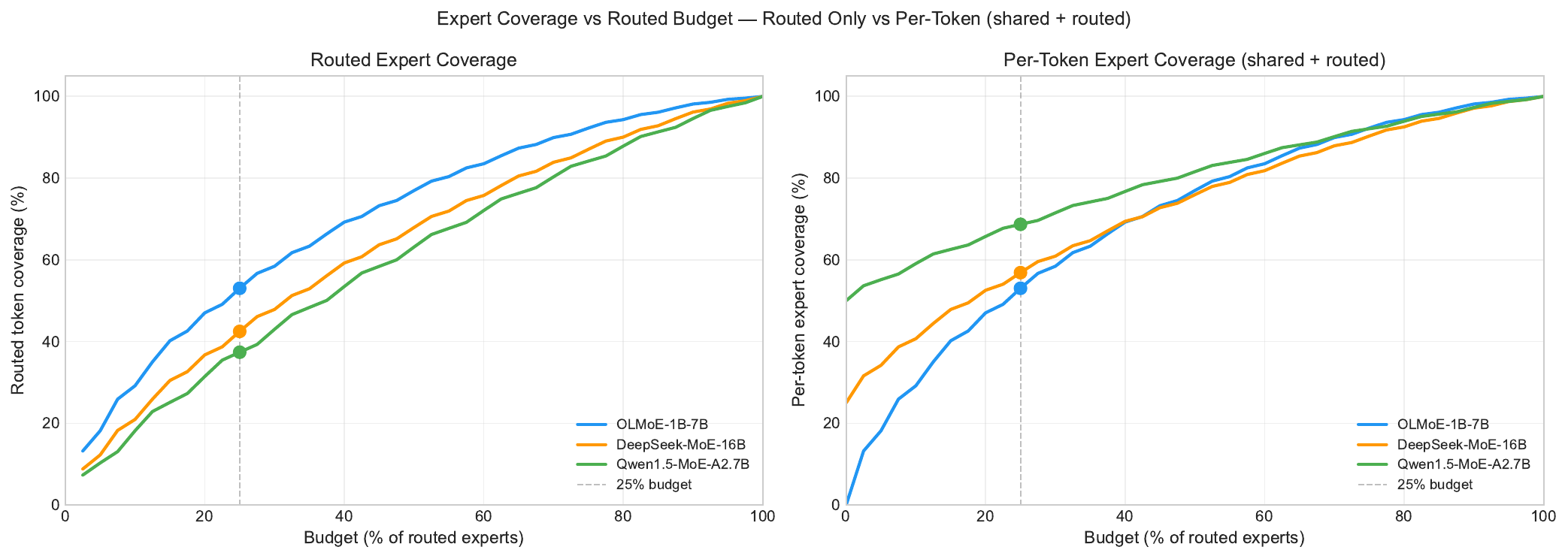}
\caption{Expert coverage as a function of routed expert budget for all three
  models. \emph{Left:} fraction of per-layer token activations captured by the
  top-$k$ routed experts. \emph{Right:} per-token expert coverage, counting both
  shared experts (always active) and the selected routed experts.
  Qwen's four shared experts provide a 50\% floor before any routed expert
  is selected; OLMoE has no shared experts and starts from 0\%.
  Dots mark the 25\% operating point.}
\label{fig:fixed-budget}
\end{figure}

\paragraph{Coverage and cold experts.}
At a 25\% \emph{routed} expert budget ($k$~=~16 for OLMoE, 15 for Qwen, 16 for DeepSeek),
the selected experts capture 53.0\%, 37.4\%, and 42.6\% of per-layer token
activations respectively---far more than the 25\% expected under uniform routing.
Counting the shared experts (always active) alongside the selected routed
experts, per-token expert coverage rises to 53\%, 69\%, and 57\% for OLMoE,
Qwen, and DeepSeek (Figure~\ref{fig:fixed-budget}; values averaged across all
10 datasets).
The complement is a cold tail: 28.5\% of OLMoE's routed experts receive less
than half the uniform share per layer, against 11.0\% for DeepSeek and 5.2\%
for Qwen---with no shared experts to absorb diffuse traffic, OLMoE concentrates
all routing through its routed pool. On Spider---a code task and, as
Figure~\ref{fig:cv-comparison} shows, among the most concentrated
datasets---60\% of activations flow to just 12.8 routed experts per layer
on average (range 9--21); broader tasks show less extreme distributions,
but a cold tail persists across all datasets.

The large variation in cold-expert fraction across models is explained by
\emph{shared-expert absorption}. Qwen's four shared experts process every token
and absorb a substantial fraction of task-agnostic capacity, leaving less
variation for the routed pool to express---hence its flatter routing
distribution and lower cold\%. DeepSeek's two shared experts produce an
intermediate effect. OLMoE, with no shared experts, pushes all specialisation
through the routed pool, producing the sharpest per-layer concentration. Once
shared capacity is accounted for, models converge: diffuse tasks (BoolQ, MMLU,
Wikitext) require nearly the same \emph{routed} expert budget across all three
architectures.

\begin{figure}[!htbp]
\centering
  \includegraphics[width=0.72\textwidth]{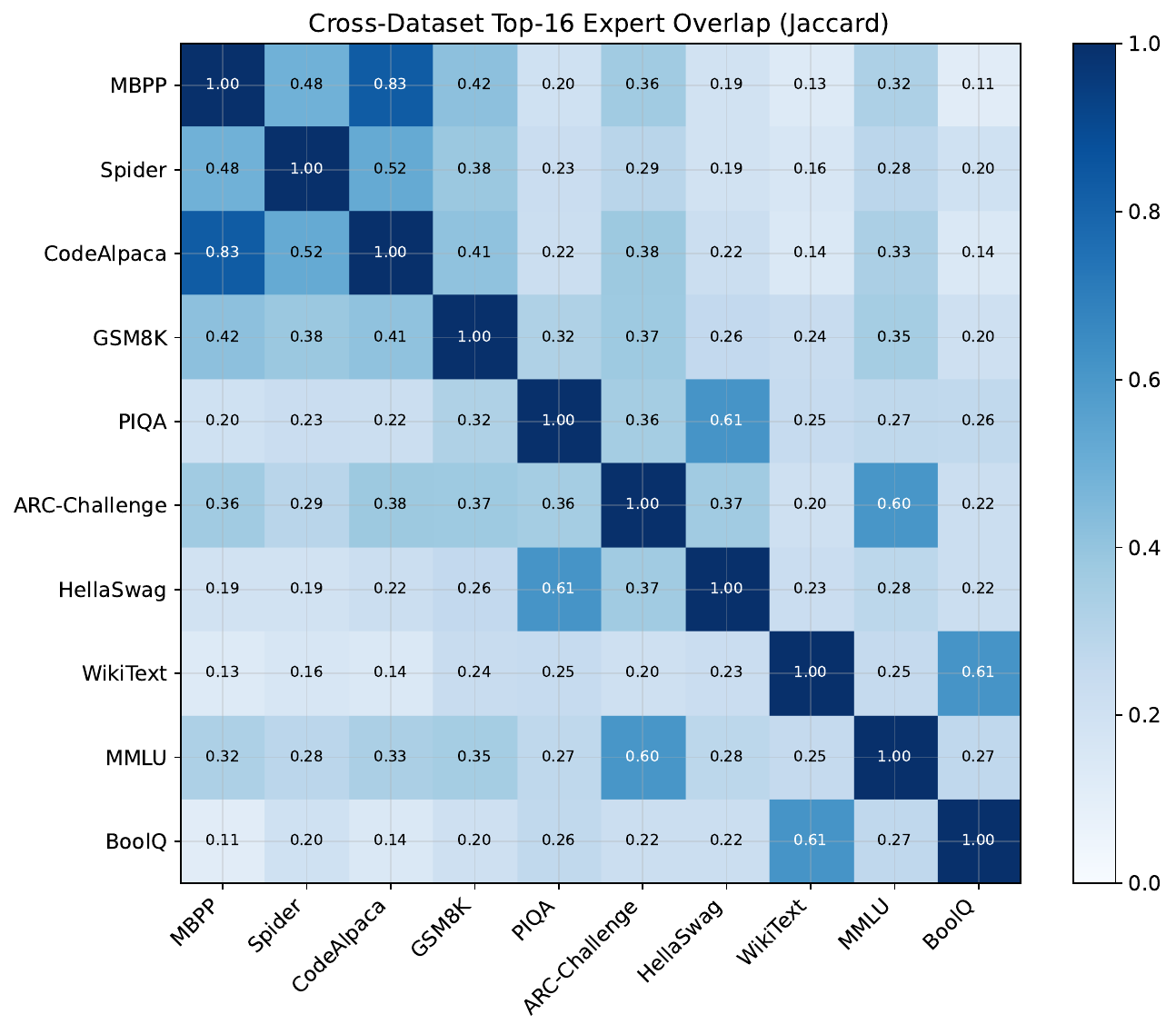}
\caption{Cross-dataset Jaccard similarity of top-16 expert sets for OLMoE.
  Code/programming tasks (MBPP, Spider, CodeAlpaca) share experts with each
  other but not with broad-domain tasks, confirming domain-driven
  task-dependent specialisation.}
\label{fig:jaccard}
\end{figure}

\paragraph{Task-dependence.}
The identity of hot experts changes with the input distribution. Code and
programming tasks (MBPP, Spider, CodeAlpaca) produce the lowest routed-expert
entropy---in OLMoE, $H \approx 0.83$--$0.86$ vs.\ $\approx 0.97$ for broad
tasks like HellaSwag or Wikitext. Cross-dataset overlap between the selected
expert sets---measured by the Jaccard index---is \emph{structured} and domain-driven: code tasks cluster tightly
(MBPP--CodeAlpaca $J = 0.83$, Spider--CodeAlpaca $J = 0.52$), reasoning tasks
form a separate cluster (ARC-Challenge--MMLU $J = 0.60$, PIQA--HellaSwag
$J = 0.61$), while cross-domain pairs are sharply dissimilar
(MBPP--Wikitext $J = 0.13$, Spider--Wikitext $J = 0.16$). This confirms that the router expresses genuine task-dependent specialisation,
not merely generic concentration on a fixed subset of experts, and justifies
profiling each task separately (Figure~\ref{fig:jaccard}). The same
domain-driven structure holds for DeepSeek and Qwen, with slightly weaker
within-domain clustering in Qwen.

\paragraph{Depth.}
Routing skew is substantially higher from mid-layers onward than in the
earliest layers. In OLMoE on Spider, layer CV rises sharply from 0.82
(layer~0) to around 1.4--1.6 from layer~4 onward (peaking at 1.64 in
layer~11), and coverage at 25\% increases from 53\% to 75\%. The same
pattern holds for DeepSeek (CV~0.53 at layer~1, peak~1.19 at layer~20)
and Qwen (CV~0.43 at layer~0, peak~0.93 at layer~20): despite differing
absolute values---lower for fine-grained architectures due to shared-expert
absorption---all three models show a consistent $\approx$2$\times$
amplification from early to peak layers,
consistent with the notion that early layers perform broader processing
while deeper layers specialise.

\subsection{Profiling is Cheap and Robust}
\label{sec:profiling-stability}

Subsample stability is high across all three architectures: profiling on a
10\% random subsample of the calibration set recovers the same top-$k$ expert
sets as the full set, with mean Jaccard~$\geq$~0.94 for all models (OLMoE:
0.979, DeepSeek: 0.962, Qwen: 0.941; 50 bootstrap trials per dataset).
MBPP (374 training examples, 10\%~$\approx$~37 samples) is the only dataset
below 0.90, and only for the fine-grained architectures (DeepSeek: 0.86,
Qwen: 0.81); for OLMoE and for every other dataset, mean stability exceeds
$J = 0.90$. Even a small data slice produces a reliable expert ranking.

A second dimension of robustness concerns the choice of profiling signal.
Experts can be ranked either by \emph{activation count} (how many tokens
were routed to each expert) or by \emph{routing mass} (the cumulative
softmax probability weight assigned to each expert). At the default 25\%
routed-expert budget, the two signals produce almost identical selections
for OLMoE (mean per-layer Jaccard~0.981, averaged across datasets) and
remain fairly close for Qwen (0.920), the two architectures used in our
fine-tuning experiments. DeepSeek diverges much more substantially (0.646),
consistent with its finer-grained experts creating a longer low-weight tail
(\S\ref{sec:abl-counts-mass}). We therefore use the simpler count signal as
the default throughout, while noting that mass-based ranking may be
preferable for DeepSeek-like architectures.

Appendix~\ref{app:profiling} provides full supporting detail: per-layer CV
and cold-expert statistics (\S\ref{app:skew}), activation heatmaps for all
10 datasets (\S\ref{app:imbalance}), and per-dataset subsample stability
charts across all three models (\S\ref{app:stability}). Together, these
results motivate a simple profiling design: count activations over a small
calibration set, rank experts independently within each layer, and adapt
only the hot subset.

% ══════════════════════════════════════════════════════════════════════
% 4. METHOD
% ══════════════════════════════════════════════════════════════════════
\section{MoE-Sieve: Method}
\label{sec:method}

The profiling study in \S\ref{sec:profiling} shows that, within each layer,
a small subset of experts handles most tokens while the majority remain
largely cold---and both aspects of this skew hold consistently across all
three model families examined. MoE-Sieve exploits this structure with a
three-step pipeline.

\paragraph{Step 1: Profile.}
Run a single forward pass over the task training data (or a small subsample).
For each MoE layer~$l$, record the activation count
$c_l(e)$---the number of tokens routed to expert~$e$.
This requires only inference: no gradient computation or optimizer state.
As shown in \S\ref{sec:profiling-stability}, it converges with as little as
10\% of the training data and produces consistent expert rankings across all
three model families examined.

\paragraph{Step 2: Select.}
Rank experts by activation count within each layer and select the top-$k$
most-routed. Our default is $k = \lfloor 0.25 \times n_\text{routed}
\rfloor$, \ie 25\% of the routed expert pool. This threshold was identified
empirically on OLMoE: the very low-$k$ regime clearly underfits, after which
performance rises and then plateaus. Individual tasks converge at different
$k$ values, but by $k = 16$ all three had reached or were within noise of
full-LoRA parity, making it a natural and conservative operating point. The
same budget transfers in our experiments to Qwen without adjustment
(\S\ref{sec:results}). Because routing skew is layer-specific
(\S\ref{sec:cross-model-profiling}), the selected set differs across layers,
yielding a sparse adapter footprint that covers the routing-active fraction
of the network.

\paragraph{Step 3: Fine-tune.}
Attach LoRA adapters \citep{hu2022lora} to the selected routed experts,
reducing trainable expert parameters by 75\% relative to full LoRA.
Attention layers, router gates, and shared experts (where present) are
always trained; the only variation across conditions is \emph{which routed
experts} receive adapters. This keeps all other trainable parameters
constant and ensures a controlled comparison. Alternative allocation
strategies (greedy budget, coverage-threshold) are evaluated in
\S\ref{sec:ablations}; uniform top-$k$ remains the recommended default.

% ══════════════════════════════════════════════════════════════════════
% 5. EXPERIMENTAL SETUP + MAIN RESULTS
% ══════════════════════════════════════════════════════════════════════
\section{Experimental Setup and Main Results}
\label{sec:results}

\subsection{Setup}
\label{sec:exp-setup}

\paragraph{Models.}
We fine-tune the two most architecturally distinct models from our profiling
study: OLMoE-1B-7B \citep{muennighoff2024olmoe} (64 routed experts, top-8,
16 MoE layers, no shared experts) and Qwen1.5-MoE-A2.7B \citep{qwen2024moe} (60 routed + 4
shared experts, top-4, 24 MoE layers).

\paragraph{Tasks.}
Spider\citep{yu2018spider} (text-to-SQL, evaluated by official Test Suite
execution accuracy),
GSM8K \citep{cobbe2021gsm8k} (grade-school math, exact-match accuracy), and
HellaSwag \citep{zellers2019hellaswag} (commonsense reasoning, normalised
accuracy). These cover structured generation, symbolic reasoning, and
commonsense understanding; Spider and GSM8K also showed the strongest
cross-model profiling divergence (\S\ref{sec:cross-model-profiling}), making
them informative stress tests.

\paragraph{Conditions.}
We compare two main conditions:
(1)~\emph{Full LoRA}---LoRA adapters on all routed experts; and
(2)~\emph{Hot-25\%}---LoRA adapters on the top-25\% most-routed
experts per layer (16/64 for OLMoE, 15/60 for Qwen), as determined by the
profiling step in \S\ref{sec:method}. Additional controls are reported in
\S\ref{sec:ablations}. In both conditions, attention, router, and
shared-expert adapters are always active.

\paragraph{Training details.}
LoRA rank~32, $\alpha$~=~64, dropout~0.05, applied to all linear projections
of selected modules. AdamW optimiser, learning rate $4 \times 10^{-4}$,
3 epochs, effective batch size~64, 8 seeds per condition.

\subsection{Main Result: 25\% of Experts Is Competitive with Full LoRA}
\label{sec:main-result}

The central question is whether selecting only the top-25\% of routed experts
per layer costs anything in accuracy. Table~\ref{tab:main} answers this
with three complementary lenses: mean accuracy with seed variance,
paired seed-level delta with 95\% CI, and formal equivalence at a
pre-declared $\pm$2~pp margin (full TOST in Appendix~\ref{app:stats}).

The short answer is: essentially nothing. All mean differences are within
$\pm$1~pp. At the $\pm$2~pp margin, 5 of 6 conditions formally establish
equivalence; the only exception is OLMoE~$\times$~Spider, where the mean
delta is a small $+$0.30~pp \emph{in favour} of hot-25\%, but full
LoRA's unusually high seed-to-seed variance (std~=~0.026) widens the CI
beyond the margin (\S\ref{sec:beyond}).
Qwen~$\times$~Spider shows the most cautious picture ($\Delta=-0.93$~pp,
CI~[$-$1.88,~$+$0.03]): the interval is nearly entirely below zero but
still technically includes it.
Beyond accuracy, hot-25\% standard deviations are equal to or lower than
full LoRA's in 5 of 6 conditions, indicating that selective expert tuning
is often at least as stable as full LoRA (\S\ref{sec:beyond}).

\begin{table}[ht!]
\centering
\caption{Main results: mean accuracy $\pm$ std (8 seeds), paired delta
  $\Delta$~=~hot$-$full (pp), paired 95\% CI, and equivalence at a
  pre-declared $\pm$2~pp margin ($\checkmark$ = established,
  $\times$ = inconclusive). Attention, router, and shared-expert adapters
  are always active.}
\label{tab:main}
\small
\begin{tabular}{llccccc}
\toprule
\textbf{Model} & \textbf{Task}
  & \textbf{Full LoRA} & \textbf{Hot (25\%)}
  & $\Delta$ \textbf{(pp)} & \textbf{95\% CI (pp)}
  & \textbf{Eqv@2pp} \\
\midrule
OLMoE & Spider    & $.396 \pm .026$ & $.399 \pm .015$ & $+0.30$ & [$-2.04,\ +2.64$] & $\times$ \\
OLMoE & GSM8K     & $.304 \pm .011$ & $.304 \pm .006$ & $-0.08$ & [$-1.45,\ +1.30$] & $\checkmark$ \\
OLMoE & HellaSwag & $.805 \pm .005$ & $.807 \pm .008$ & $+0.17$ & [$-0.71,\ +1.05$] & $\checkmark$ \\
\midrule
Qwen  & Spider    & $.520 \pm .014$ & $.511 \pm .005$ & $-0.93$ & [$-1.88,\ +0.03$] & $\checkmark$ \\
Qwen  & GSM8K     & $.590 \pm .011$ & $.592 \pm .007$ & $+0.20$ & [$-0.77,\ +1.17$] & $\checkmark$ \\
Qwen  & HellaSwag & $.885 \pm .002$ & $.893 \pm .001$ & $+0.73$ & [$+0.53,\ +0.93$] & $\checkmark$ \\
\bottomrule
\end{tabular}
\normalsize
\end{table}

The accuracy parity comes with a substantial reduction in training cost
(Table~\ref{tab:efficiency}). Hot-25\% reduces total LoRA trainable
parameters by 70--73\% and adapter checkpoint size by 71--73\%. Wall-clock
training time is reduced by up to $\sim$50\%, though the exact speedup
varies by task and hardware configuration. Note that because attention,
router, and shared-expert adapters are always active, the savings reflect
only the routed-expert reduction.

\begin{table}[ht!]
\centering
\caption{Efficiency of hot-25\% relative to full LoRA. Trainable-parameter
  counts are from the final PEFT model. Wall-clock times are one example
  measured on GSM8K (3 epochs, 351 steps) on a single GPU; actual speedup
  varies by task and configuration.}
\label{tab:efficiency}
\small
\begin{tabular}{lccccc}
\toprule
\textbf{Model} & \textbf{Params (Full $\rightarrow$ Hot)} & \textbf{Red.}
  & \textbf{Ckpt (Full $\rightarrow$ Hot)} & \textbf{Red.}
  & \textbf{Time (Full $\rightarrow$ Hot)} \\
\midrule
OLMoE & 311.5M $\rightarrow$ 85.0M & 72.7\%
  & 1.25\,GB $\rightarrow$ 340\,MB & 73.4\%
  & 1h\,48m $\rightarrow$ 54m\,(50\%) \\
Qwen  & 509.7M $\rightarrow$ 151.3M & 70.3\%
  & 2.04\,GB $\rightarrow$ 606\,MB & 71.0\%
  & 3h\,23m $\rightarrow$ 1h\,44m\,(49\%) \\
\bottomrule
\end{tabular}
\normalsize
\end{table}

The next two sections probe further: \S\ref{sec:ablations} tests whether
the routing signal is necessary and whether smarter allocation strategies
improve on the uniform rule; \S\ref{sec:beyond} examines secondary patterns
in stability and budget dynamics.

% ══════════════════════════════════════════════════════════════════════
% 6. PRACTICAL ABLATIONS / CONTROLS
% ══════════════════════════════════════════════════════════════════════
\section{Practical Ablations}
\label{sec:ablations}

The main result in \S\ref{sec:main-result} shows that selecting the
top-25\% most-routed experts per layer is competitive with full LoRA at a
fraction of the parameter cost. This section tests the assumptions behind
that result. Is the routing signal actually necessary, or would \emph{any}
subset of experts work equally well? Does a smarter, non-uniform allocation
of experts across layers outperform the simple uniform rule? And does
ranking experts by raw activation counts versus gate-weighted mass change
which experts are selected?

\subsection{Random Baseline}
\label{sec:abl-random}

To isolate the contribution of routing-informed selection from simple
parameter reduction, we compare hot-$k$ and random-$k$ at matched budget
on OLMoE~$\times$~GSM8K. We test two budget levels: $k$~=~16, our main operating point (25\% of
64 routed experts), and $k$~=~8, OLMoE's per-token routing width---the minimum budget where it
is at least theoretically possible to cover every token's full routing
path with adapters. Random-$k$ selects $k$ experts per
layer uniformly at random:

\begin{itemize}
  \item $k$~=~16: hot $0.304 \pm 0.006$ vs random $0.279 \pm 0.008$
    ($\Delta = +2.5$~pp)
  \item $k$~=~8:\phantom{0} hot $0.291 \pm 0.007$ vs random $0.270 \pm 0.011$
    ($\Delta = +2.1$~pp)
\end{itemize}

\noindent
At both budgets, routing-guided selection outperforms random by
2--2.5~pp. Random-$k$ also shows higher variance at $k$~=~8
(std~=~0.011 vs 0.007), consistent with an uninformed selection being
more sensitive to initialisation under a tight budget. Strikingly,
random selection at $k$~=~16 (0.279) underperforms even hot $k$~=~8
(0.291): doubling the adapter budget with the wrong experts yields worse
results than half the budget with the right ones. This confirms that the
profiling signal captures genuine task-specific specialisation,
consistent with the cross-dataset Jaccard clustering in
\S\ref{sec:cross-model-profiling}.

\subsection{Dynamic Allocation Strategies}
\label{sec:abl-budget}

The profiling study in \S\ref{sec:profiling} shows that routing skew
varies across layers. A natural follow-up is whether the expert budget
should also vary---allocating fewer experts in concentrated layers and
more in balanced ones where coverage requires a larger subset. We test
two such \emph{dynamic} strategies.

\paragraph{Greedy marginal-gain allocation.}
Given the same total expert--layer slots as uniform top-$k$, this
strategy assigns each slot to the layer--expert pair that maximises
cumulative coverage gain. Under concave coverage gains---which hold in
practice---this makes greedy allocation a natural upper-bound-oriented
baseline and produces a non-uniform
per-layer allocation (Appendix~\ref{app:greedy}, Table~\ref{tab:greedy-uniform}).
Table~\ref{tab:greedy} compares all six conditions.

\begin{table}[ht!]
\centering
\caption{Greedy marginal-gain allocation vs.\ uniform hot-25\%.
  Both strategies use the \emph{same} total expert--layer budget;
  greedy varies the per-layer count ($k$ range) while uniform fixes it.
  Mean routing coverage (\%) is virtually identical, and accuracy
  differences are negligible across all conditions (8 seeds each).}
\label{tab:greedy}
\small
\begin{tabular}{llcccc}
\toprule
\textbf{Model} & \textbf{Task}
  & \textbf{Uniform} & \textbf{Greedy}
  & \textbf{$k$ range} & \textbf{Cov.\,(\%)} \\
\midrule
OLMoE & GSM8K     & $.304 \pm .006$ & $.304 \pm .008$ & 12--22 & 56.8 \\
OLMoE & Spider    & $.399 \pm .015$ & $.405 \pm .019$ & 12--20 & 68.4 \\
OLMoE & HellaSwag & $.807 \pm .008$ & $.806 \pm .008$ &\phantom{0}9--21 & 43.0 \\
\midrule
Qwen  & GSM8K     & $.592 \pm .007$ & $.591 \pm .010$ & 12--19 & 34.2 \\
Qwen  & Spider    & $.511 \pm .005$ & $.509 \pm .012$ & 12--19 & 48.2 \\
Qwen  & HellaSwag & $.893 \pm .001$ & $.895 \pm .001$ & 11--20 & 36.4 \\
\bottomrule
\end{tabular}
\end{table}

No significant accuracy difference emerges in any condition, despite
greedy's coverage optimality. This suggests that routing coverage is a
useful but not sufficient proxy for fine-tuning utility: uniform top-$k$
already selects the experts that matter most for gradient flow, and
redistributing slots toward balanced layers adds coverage on experts that
contribute little to task performance.

\paragraph{Coverage-threshold allocation.}
\label{sec:abl-cov}
Instead of fixing a global $k$, this strategy selects, per layer, the
minimum number of experts needed to capture a target fraction of routing
mass (we test 60\%). On OLMoE, cov60 achieves comparable accuracy to
hot-25\% (GSM8K: 0.300 vs 0.304; HellaSwag: 0.811 vs 0.807). However,
the 60\% threshold results in more total expert--layer slots than
uniform-25\%, so the comparison is \emph{confounded by parameter count}:
any advantage could reflect having more adapted parameters rather than
smarter allocation.

\medskip
\noindent
Neither dynamic strategy improves over uniform top-$k$, which we
therefore recommend as the default. Both use the same profiling data
already collected in Step~1, so the added computation cost is negligible.
Dynamic allocation may prove beneficial for architectures with greater
layer-to-layer variation in routing skew than the models studied here.

\subsection{Counts vs.\ Mass Ranking}
\label{sec:abl-counts-mass}

The profiling step can rank experts by \emph{routing counts}---how many
tokens are routed to each expert---or by \emph{routing mass}---the sum of
gate weights across all tokens, reflecting both frequency and confidence
of selection. For hard top-$k$ routing the two are correlated but not
identical: mass up-weights experts selected with high gate confidence,
while counts treat all selections equally regardless of weight. In
practice, this means an expert can be selected frequently as a low-weight
secondary route and therefore rank high by count but not by mass.

The relevant comparison is whether the two rankings select the same
top-$k$ expert set at our operating point: 25\% of routed experts per
layer.
For OLMoE, count- and mass-ranked top-$k$ sets are almost identical
(mean per-layer Jaccard~0.981, averaged across datasets), making the
choice of signal effectively inconsequential. Qwen shows a modest but
still limited divergence (mean Jaccard~0.920). DeepSeek differs much
more substantially: at the same 25\% budget, the count- and mass-ranked
top-$k$ sets have mean Jaccard~0.646.

We interpret this pattern as being consistent with
DeepSeek's finer-grained experts producing a longer tail of low-weight
selections, which inflates counts relative to mass. This interpretation is
plausible but not directly established by the present study.

We use count-based ranking as the default for OLMoE and Qwen: under hard
top-$k$ routing, the router makes a binary selection decision, making
activation count the simplest and most interpretable first-order signal.
For DeepSeek-like architectures, however, mass-based ranking may better
reflect which experts receive meaningful routing weight. Whether that
distinction translates into a meaningful fine-tuning difference remains
open, since DeepSeek was not fine-tuned in this study.

% ══════════════════════════════════════════════════════════════════════
% 7. BEYOND THE MAIN RESULT
% ══════════════════════════════════════════════════════════════════════
\section{Beyond the Main Result: Stability and Budget Dynamics}
\label{sec:beyond}

The accuracy comparison in \S\ref{sec:main-result} is the paper's central
claim. Beyond that fixed hot-25\% comparison, the experiments reveal two
secondary behaviours of MoE-Sieve---one about stability across seeds and one
about how performance evolves with expert budget---followed by a tentative
interpretation linking both.

\subsection{Seed Stability Under MoE-Sieve}
\label{sec:var-reduction}

Table~\ref{tab:variance} reports seed-to-seed standard deviation for each
model--task--condition pair. At the 25\% working point, MoE-Sieve reduces
standard deviation relative to full LoRA in 5 of 6 conditions. The reductions
are substantial on structured prediction and arithmetic: 43\% for OLMoE and
64\% for Qwen on Spider, and 41--42\% on GSM8K for both models. On Qwen,
MoE-Sieve reduces variance across all three tasks.

\begin{table}[ht!]
\centering
\caption{Seed-to-seed standard deviation (8 seeds). \emph{Ratio}~=
  std(hot-25\%) / std(full LoRA); values $<$1 indicate variance reduction.}
\label{tab:variance}
\begin{tabular}{llccc}
\toprule
\textbf{Model} & \textbf{Task} & \textbf{Full} & \textbf{Hot (25\%)} & \textbf{Hot / Full} \\
\midrule
OLMoE & Spider    & .026 & .015 & 0.57$\times$ \\
OLMoE & GSM8K     & .011 & .006 & 0.58$\times$ \\
OLMoE & HellaSwag & .005 & .008 & 1.68$\times$ \\
\midrule
Qwen  & Spider    & .014 & .005 & 0.36$\times$ \\
Qwen  & GSM8K     & .011 & .007 & 0.59$\times$ \\
Qwen  & HellaSwag & .002 & .001 & 0.67$\times$ \\
\bottomrule
\end{tabular}
\end{table}

The one exception is OLMoE~$\times$~HellaSwag, where hot-25\% shows slightly
higher variance (std 0.008 vs.\ 0.005); the absolute values are small and this
is the only case where the pattern reverses. For OLMoE~$\times$~Spider,
full LoRA has the highest variance of any condition (std~=~0.026). This is
what widens the paired confidence interval in \S\ref{sec:main-result} beyond
the $\pm$2pp equivalence margin, even though the mean delta is slightly
positive.

\subsection{Budget Dynamics Beyond the 25\% Working Point}
\label{sec:ushape}

\begin{figure}[ht!]
  \centering
    \includegraphics[width=0.95\textwidth]{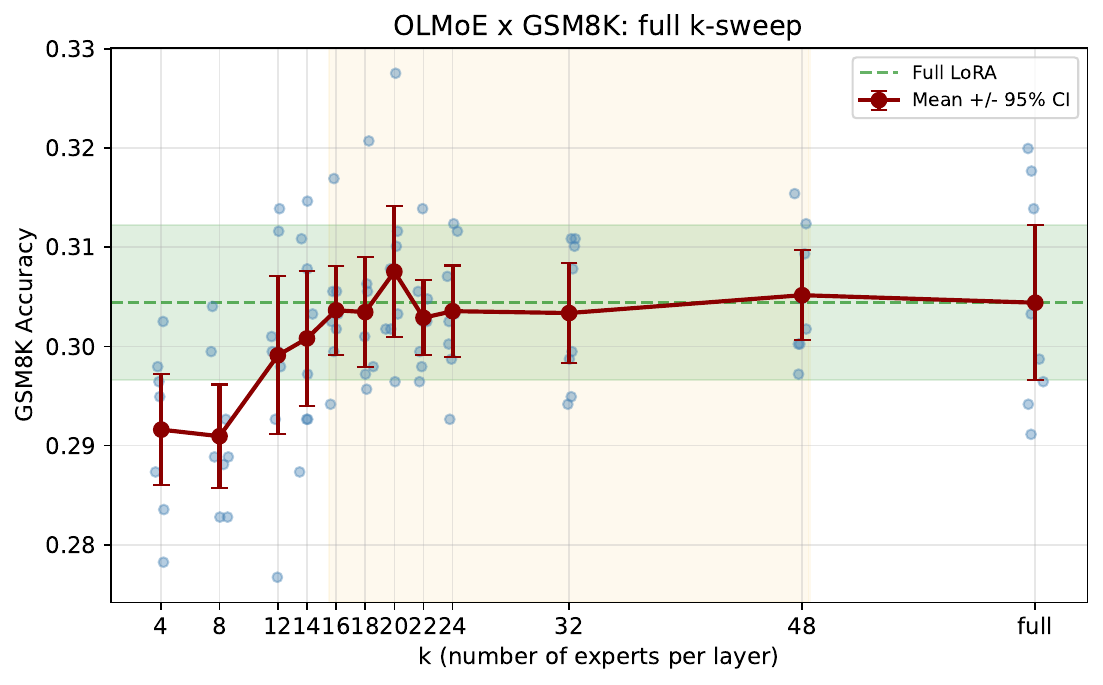}
  \caption{OLMoE $\times$ GSM8K $k$-sweep (8 seeds per $k$). Dots show
    individual seeds; red markers show mean $\pm$ 95\% CI; the green band marks
    full LoRA.}
  \label{fig:ksweep-gsm8k}
  \end{figure}

Moving beyond the fixed hot-25\% setting, Figure~\ref{fig:ksweep-gsm8k}
shows the full $k$-sweep on OLMoE~$\times$~GSM8K. Mean accuracy reaches the
full-LoRA level by $k$~=~16 and then plateaus: several intermediate budgets
match or slightly exceed the full-LoRA mean while using a fraction of the
routed-expert adapter parameters. The densest endpoint is therefore not
obviously the best operating point for MoE-Sieve.

A similar pattern appears on OLMoE~$\times$~HellaSwag: $k$~=~32 slightly
exceeds full LoRA accuracy (.812 vs.\ .805; Appendix~\ref{app:ksweep},
Table~\ref{tab:ksweep-supp}). These gains are small and observed on a single
model, so we treat them as suggestive rather than definitive. Still, they
point in the same direction: once the hot experts are covered, adding colder
experts may not improve mean accuracy and can coincide with worse stability.

\subsection{A Cold-Expert Noise Hypothesis}
\label{sec:cold-noise}

One interpretation consistent with the two MoE-Sieve behaviours above is that
rarely-activated experts act as noise sources during fine-tuning. Cold experts
receive sparse and inconsistent gradient updates, making their adapter weights
sensitive to random initialisation and data ordering. Including them, as full
LoRA does, may inflate seed-to-seed variance without contributing to
accuracy---which would explain why performance saturates before $k$~=~64 and
why several intermediate budgets are more stable.

This view is also consistent with the random-baseline ablation
(\S\ref{sec:abl-random}): routing-guided selection outperforms random selection
at matched budget by 2--2.5~pp. A randomly chosen expert set is more likely
to include cold experts, which is directionally consistent with the observed
gap.

We treat this as a hypothesis, not a causal claim. OLMoE~$\times$~HellaSwag
does not show variance reduction at hot-25\%, and the $k$-sweep evidence comes
from a single model--task pair with 8 seeds. A direct test---hot experts
supplemented with randomly-selected cold fillers at the same total
budget---would isolate the effect of cold-expert inclusion from budget size,
and is left for future work.

% ══════════════════════════════════════════════════════════════════════
% 8. CONCLUSION, LIMITATIONS, AND FUTURE WORK
% ══════════════════════════════════════════════════════════════════════
\section{Conclusion, Limitations, and Future Work}
\label{sec:conclusion}

This paper studies routing-guided expert selection for parameter-efficient
fine-tuning of Mixture-of-Experts models. Across three architectures and ten
datasets, we show that global load balancing can hide substantial
layer-local routing skew: within a given layer, a relatively small subset of
experts carries most of the traffic. This makes uniform expert adaptation a
poor match to the routing patterns actually seen at fine-tuning time.

In our experiments, restricting LoRA to the top-25\% most-routed experts per
layer remains competitive with full LoRA across two models and three tasks,
while reducing LoRA trainable parameters by 70--73\%, adapter checkpoint size
by 71--73\%, and wall-clock training time by up to 50\%. Ablations further
show that the routing signal matters: random expert selection at the same
budget is 2--2.5~pp worse, while more elaborate budget-allocation schemes do
not improve over uniform top-$k$.

Beyond efficiency, the budget sweeps indicate that full LoRA is not always the
best operating point. Once the hot experts are covered, adding colder experts
often brings little benefit and can coincide with higher seed-to-seed
variability. We treat the cold-expert noise account as a tentative
interpretation of this pattern rather than a causal claim.

The practical takeaway is simple: one profiling pass, uniform per-layer
top-$k$ at 25\%, then standard LoRA training.

\paragraph{Limitations.}
Our fine-tuning experiments cover two MoE models, OLMoE-1B-7B and
Qwen1.5-MoE-A2.7B, with roughly 7B and 14B total parameters
respectively; larger MoE models remain untested, and the 25\% threshold
may shift at larger scale. Task coverage is limited to three domains
(structured generation, math reasoning, commonsense); safety alignment,
instruction following, and multilingual settings are not evaluated.
Expert selection is static---determined once before training---and
dynamic re-profiling during training could capture shifting routing
patterns but adds complexity.

\paragraph{Future work.}
Several directions emerge from this study. A formal account linking the
routing Pareto distribution to the fine-tuning capacity curve could
replace the empirical 25\% threshold with a principled selection
criterion. The cold-expert noise hypothesis awaits a controlled
experiment that would isolate the causal mechanism. Dynamic allocation
strategies, which showed comparable results to uniform top-$k$ in our
experiments, may become beneficial for architectures with greater
layer-to-layer variation in routing skew. Finally, scaling the study to
larger MoE models and more diverse tasks would establish the generality
of the findings.

% ══════════════════════════════════════════════════════════════════════
% REFERENCES
% ══════════════════════════════════════════════════════════════════════
\bibliographystyle{plainnat}
\bibliography{references}

% ══════════════════════════════════════════════════════════════════════
% APPENDIX
% ══════════════════════════════════════════════════════════════════════
\clearpage
\appendix

\section{Profiling Details}
\label{app:profiling}

% ── A.1 ──────────────────────────────────────────────────────────────
\subsection{Per-Layer Expert Utilization}
\label{app:skew}

Table~\ref{tab:perlayer-spider} reports per-layer routing statistics for
OLMoE on Spider. CV and cold-expert fraction rise sharply from layer~4
onward, peaking at layer~11 (CV~=~1.63, Top-16 coverage~=~75\%); the first
three layers are markedly more balanced. The cross-model depth pattern and
the 4.0--4.9$\times$ global-to-layer ratio across all 30 model--dataset
combinations are shown in Figure~\ref{fig:cv-comparison} (main text, §\ref{sec:cv}).

\begin{table}[H]
\centering
\footnotesize
\begin{tabular}{rcccc}
\toprule
\textbf{Layer} & \textbf{CV} & \textbf{Cold\%} & \textbf{Top-16 Cov\%} & \textbf{Norm. Entropy} \\
\midrule
 0 & 0.81 & 25\% & 53\% & 0.935 \\
 1 & 0.85 & 23\% & 52\% & 0.938 \\
 2 & 1.08 & 41\% & 59\% & 0.902 \\
 3 & 1.09 & 47\% & 64\% & 0.886 \\
 4 & 1.38 & 55\% & 71\% & 0.845 \\
 5 & 1.46 & 47\% & 70\% & 0.837 \\
 6 & 1.42 & 50\% & 72\% & 0.828 \\
 7 & 1.54 & 50\% & 73\% & 0.815 \\
 8 & 1.50 & 58\% & 73\% & 0.819 \\
 9 & 1.43 & 53\% & 71\% & 0.836 \\
10 & 1.30 & 50\% & 68\% & 0.857 \\
11 & 1.63 & 55\% & 75\% & 0.809 \\
12 & 1.41 & 58\% & 72\% & 0.838 \\
13 & 1.40 & 56\% & 73\% & 0.835 \\
14 & 1.37 & 55\% & 72\% & 0.841 \\
15 & 1.53 & 59\% & 74\% & 0.818 \\
\bottomrule
\end{tabular}
\caption{Per-layer profiling for OLMoE on Spider. CV and cold-expert
  fraction rise sharply from layer~4 onward. Cold\% = fraction of experts
  receiving $<$50\% of the uniform routing share.}
\label{tab:perlayer-spider}
\end{table}

% ── A.2 ──────────────────────────────────────────────────────────────
\subsection{Structure of the Imbalance}
\label{app:imbalance}

\begin{figure}[ht!]
  \centering
  \includegraphics[width=\textwidth]{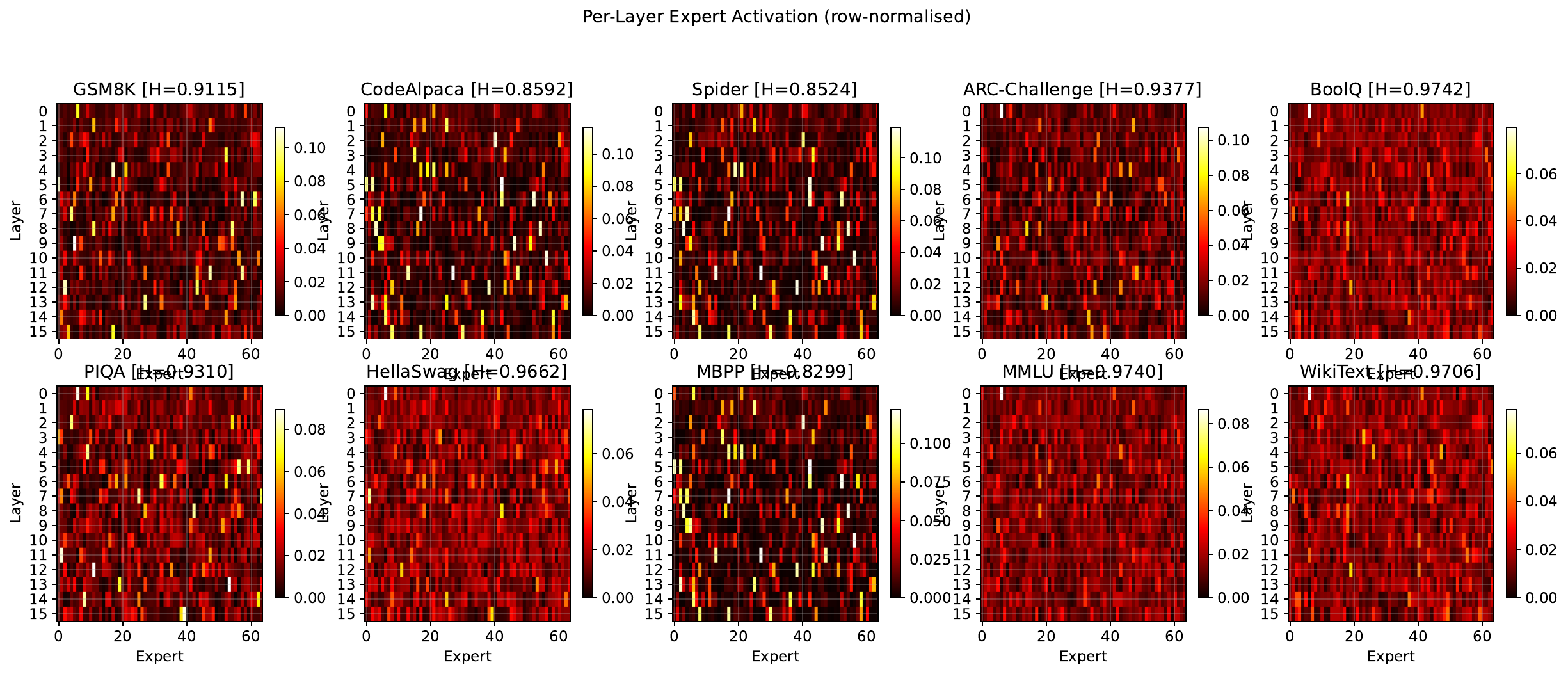}
  \caption{Per-layer expert activation heatmaps (row-normalised) for OLMoE
    across all 10 profiling datasets. Each row is a layer and each column an
    expert. All datasets exhibit a hot-head plus cold-tail structure, but its
    sharpness varies with task domain: code and programming tasks (MBPP,
    Spider, CodeAlpaca) show more concentrated patterns than broad-domain
    tasks (Wikitext, BoolQ, MMLU).}
  \label{fig:heatmaps}
\end{figure}

The per-token expert coverage curves (routed + shared) are in
Figure~\ref{fig:fixed-budget} (§\ref{sec:cross-model-profiling}).
Figure~\ref{fig:heatmaps} shows activation heatmaps for all 10 datasets,
illustrating how the concentration sharpness varies by task domain.

% ── A.3 ──────────────────────────────────────────────────────────────
\subsection{Cheap and Robust Profiling}
\label{app:stability}

\begin{figure}[ht!]
\centering
  \includegraphics[width=\textwidth]{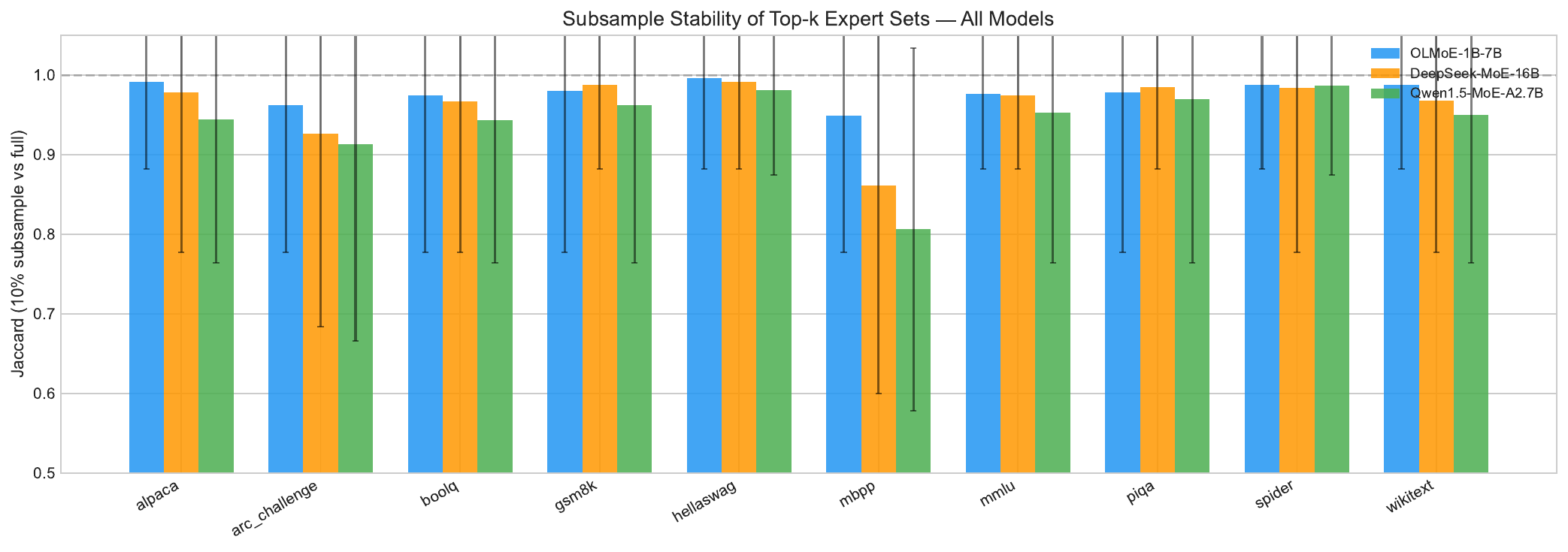}
\caption{Subsample stability of top-$k$ expert sets under 10\% profiling
  data, for all three models across all datasets. Bars show mean Jaccard
  across 50 bootstrap trials; error bars show the gap to the minimum
  per-layer overlap. MBPP (374 examples) is the only dataset with mean
  $J < 0.90$ for fine-grained architectures; all other datasets
  achieve mean $J \geq 0.90$ across all models.}
\label{fig:stability}
\end{figure}

Figure~\ref{fig:stability} shows per-dataset and per-model breakdown of
subsample stability (10\% bootstrap, 50 trials). MBPP is the only outlier,
driven by its small size (374 examples); all other datasets exceed
$J = 0.90$ for all three models.

\section{Statistical Tests}
\label{app:stats}

Table~\ref{tab:tost-full} reports TOST equivalence test $p$-values at three
margins ($\varepsilon \in \{1, 2, 3\}$ percentage points). The test evaluates
$H_0\colon |\mu_{\text{hot}} - \mu_{\text{full}}| \geq \varepsilon$ against
$H_1\colon |\mu_{\text{hot}} - \mu_{\text{full}}| < \varepsilon$, using
paired differences across 8 seeds ($\alpha$~=~0.05) \citep{schuirmann1987tost}.

\begin{table}[H]
\centering
\footnotesize
\caption{TOST equivalence $p$-values and paired $t$-test results for
  hot-25\% vs.\ full LoRA.  \checkmark = equivalence established at
  $\alpha$~=~0.05.}
\label{tab:tost-full}
\begin{tabular}{lcc rrr r}
\toprule
\textbf{Model} & \textbf{Task} & \textbf{$\Delta$ (pp)}
  & \textbf{$\varepsilon$=1pp} & \textbf{$\varepsilon$=2pp}
  & \textbf{$\varepsilon$=3pp} & \textbf{$t$-test $p$} \\
\midrule
OLMoE & GSM8K     & $-$0.08 & .003~\checkmark & $<$.001~\checkmark & $<$.001~\checkmark & .753 \\
OLMoE & HellaSwag & $+$0.17 & $<$.001~\checkmark & $<$.001~\checkmark & $<$.001~\checkmark & .243 \\
OLMoE & Spider    & $+$0.30 & .251 & .065 & .015~\checkmark & .771 \\
\midrule
Qwen  & GSM8K     & $+$0.20 & .006~\checkmark & $<$.001~\checkmark & $<$.001~\checkmark & .425 \\
Qwen  & HellaSwag & $+$0.73 & $<$.001~\checkmark & $<$.001~\checkmark & $<$.001~\checkmark & $<$.001 \\
Qwen  & Spider    & $-$0.93 & .429 & .016~\checkmark & $<$.001~\checkmark & .055 \\
\bottomrule
\end{tabular}
\end{table}

\paragraph{Summary.}
At $\varepsilon$~=~2pp, 5 of 6 conditions establish equivalence.
The only failure is OLMoE~$\times$~Spider: the mean difference is
$+$0.30~pp in favour of hot-25\%, but full LoRA's standard deviation on
this condition (0.026) is the highest across all model--task pairs
(Table~\ref{tab:variance}), producing a confidence interval
[$-$2.04,~$+$2.64] too wide for the 2pp margin. The failure is a
consequence of the reference condition's elevated variance, not of
hot-25\%'s underperformance---an observation consistent with the variance
analysis in \S\ref{sec:beyond}. At $\varepsilon$~=~3pp, all 6
conditions pass. With only 8 paired seeds, near-zero differences are
naturally harder to certify tightly than large ones; for that reason, we
emphasise paired confidence intervals and TOST equivalence rather than
superiority testing alone.

\section{Greedy Budget Allocation Algorithm}
\label{app:greedy}

The greedy marginal-gain algorithm operates as follows:

\begin{enumerate}
  \item \textbf{Input}: per-layer activation count vectors
    $\{c_l\}_{l=1}^L$, total budget $B$ (number of expert--layer slots).
  \item Initialise each layer with $k_l = 0$ selected experts.
  \item Repeat $B$ times:
    \begin{enumerate}
      \item For each layer $l$, compute the marginal coverage gain of adding
        the next-highest expert: $\Delta_l = \text{cov}(k_l + 1) - \text{cov}(k_l)$,
        where $\text{cov}(k)$ is the fraction of layer-$l$ routing mass
        captured by the top-$k$ experts.
      \item Allocate one slot to $l^* = \arg\max_l \Delta_l$.
      \item Update $k_{l^*} \leftarrow k_{l^*} + 1$.
    \end{enumerate}
  \item \textbf{Output}: per-layer expert counts $\{k_l\}$ and corresponding
    expert identity sets.
\end{enumerate}

\noindent
This is optimal for concave coverage functions (coverage gain is
non-increasing in $k$), which holds in practice. Despite this optimality
guarantee, the resulting allocation does not outperform uniform top-$k$
(\S\ref{sec:abl-budget}).

Table~\ref{tab:greedy-uniform} compares per-layer budgets for OLMoE on
GSM8K ($B$~=~256 total expert-layer slots, equal to $16\times16$).

\begin{table}[H]
\centering
\footnotesize
\caption{Per-layer allocation: uniform ($k$~=~16) vs.\ greedy for
  OLMoE~$\times$~GSM8K.  Despite greedy's variable allocation
  (12--22 experts/layer), mean coverage is virtually identical (0.567 vs.\
  0.568), explaining why downstream accuracy does not differ.}
\label{tab:greedy-uniform}
\begin{tabular}{r cc cc}
\toprule
 & \multicolumn{2}{c}{\textbf{Uniform}} & \multicolumn{2}{c}{\textbf{Greedy}} \\
\cmidrule(lr){2-3}\cmidrule(lr){4-5}
\textbf{Layer} & $k$ & Cov\% & $k$ & Cov\% \\
\midrule
 0 & 16 & 48.5 & 15 & 46.8 \\
 1 & 16 & 48.0 & 14 & 44.5 \\
 2 & 16 & 52.8 & 16 & 52.8 \\
 3 & 16 & 52.4 & 19 & 57.8 \\
 4 & 16 & 57.5 & 16 & 57.5 \\
 5 & 16 & 58.7 & 15 & 56.9 \\
 6 & 16 & 66.6 & 17 & 68.4 \\
 7 & 16 & 58.2 & 17 & 60.0 \\
 8 & 16 & 60.5 & 15 & 58.8 \\
 9 & 16 & 58.8 & 12 & 51.9 \\
10 & 16 & 55.4 & 15 & 53.6 \\
11 & 16 & 62.5 & 12 & 55.7 \\
12 & 16 & 60.3 & 15 & 58.6 \\
13 & 16 & 59.5 & 16 & 59.5 \\
14 & 16 & 51.0 & 20 & 58.3 \\
15 & 16 & 56.1 & 22 & 67.7 \\
\midrule
\textbf{Mean} & \textbf{16} & \textbf{56.7} & \textbf{16} & \textbf{56.8} \\
\bottomrule
\end{tabular}
\end{table}

\section{Supplementary Budget Sweeps}
\label{app:ksweep}

Section~\ref{sec:ushape} reports the OLMoE~$\times$~GSM8K $k$-sweep.
Below we provide supplementary HellaSwag and Spider budget sweeps for
completeness. Spider uses official Test Suite execution accuracy; HellaSwag
uses validation accuracy.

\begin{table}[H]
\centering
\small
\begin{minipage}[t]{0.47\textwidth}
\centering
\textbf{OLMoE $\times$ HellaSwag}\\[2pt]
\begin{tabular}{lc}
\toprule
\textbf{Setting} & \textbf{Accuracy} \\
\midrule
hot $k$~=~8   & $.790 \pm .011$ \\
hot $k$~=~16  & $.807 \pm .008$ \\
hot $k$~=~24  & $.806 \pm .010$ \\
hot $k$~=~32  & $\mathbf{.812} \pm .007$ \\
budget-25     & $.806 \pm .008$ \\
full LoRA     & $.805 \pm .005$ \\
\bottomrule
\end{tabular}
\end{minipage}
\hfill
\begin{minipage}[t]{0.47\textwidth}
\centering
\textbf{OLMoE $\times$ Spider}\\[2pt]
\begin{tabular}{lc}
\toprule
\textbf{Setting} & \textbf{Accuracy} \\
\midrule
hot $k$~=~8   & $.365 \pm .012$ \\
hot $k$~=~16  & $.399 \pm .016$ \\
budget-25     & $\mathbf{.405} \pm .019$ \\
full LoRA     & $.396 \pm .028$ \\
\bottomrule
\end{tabular}
\end{minipage}
\normalsize
\caption{Supplementary OLMoE budget sweeps for HellaSwag and Spider (8 seeds
  each). On HellaSwag, performance plateaus from $k$~=~16 onward and the
  budget-25 control remains in the same band as full LoRA. On Spider,
  $k$~=~8 underfits, while $k$~=~16 and budget-25 remain competitive with full
  LoRA.}
\label{tab:ksweep-supp}
\end{table}

\end{document}